\documentclass[runningheads]{llncs}
\usepackage{graphicx}
\usepackage{tikz}
\usepackage{comment}
\usepackage{amsmath,amssymb} % define this before the line numbering.
\usepackage{color}

\usepackage{amsmath}
\usepackage{amssymb}
\usepackage{multirow}
\usepackage{subcaption}

\newcommand{\etal}{\textit{et al. }}
\newcommand{\eg}{\textit{e.g.}}
\newcommand{\ie}{\textit{i.e.}}

\newcommand\norm[1]{\left\lVert#1\right\rVert}

\begin{document}
% \linenumbers
\pagestyle{headings}
\mainmatter
\def\ECCVSubNumber{7174}

\title{Panoramic Vision Transformer for Saliency Detection in 360$^\circ$ Videos}

\begin{comment}
\titlerunning{ECCV-22 submission ID \ECCVSubNumber} 
\authorrunning{ECCV-22 submission ID \ECCVSubNumber} 
\author{Anonymous ECCV submission}
\institute{Paper ID \ECCVSubNumber}
\end{comment}
%******************

%\begin{comment}
\titlerunning{Panoramic Vision Transformer}
\author{Heeseung Yun \and
Sehun Lee \and
Gunhee Kim}%\orcidID{2222--3333-4444-5555}}
\authorrunning{H. Yun et al.}
\institute{Seoul National University, Seoul, Korea \\
\email{\{heeseung.yun, shlee\}@vision.snu.ac.kr, gunhee@snu.ac.kr} \\
\url{https://github.com/hs-yn/PAVER}}
%\end{comment}
%******************
\maketitle

\begin{abstract}
  360$^\circ$ video saliency detection is one of the challenging benchmarks for 360$^\circ$ video understanding
  since non-negligible distortion and discontinuity occur in the projection of any format of 360$^\circ$ videos, and
  capture-worthy viewpoint in the omnidirectional sphere is ambiguous by nature. 
  We present a new framework named \textbf{Pa}noramic \textbf{V}ision Transform\textbf{er} (PAVER).
  We design the encoder using Vision Transformer %~\cite{dosovitskiy2020image} 
  with deformable convolution, which enables us not only to plug pretrained models from normal videos into our architecture without additional modules or finetuning but also to perform geometric approximation only once, unlike previous deep CNN-based approaches.
  Thanks to its powerful encoder, PAVER can learn the saliency from three simple relative relations among local patch features, outperforming state-of-the-art models for the Wild360 benchmark %\cite{cheng2018cube}
  by large margins without supervision or auxiliary information like class activation.
  We demonstrate the utility of our saliency prediction model with the omnidirectional video quality assessment task in VQA-ODV%~\cite{li2018bridge}
  , where we consistently improve performance without any form of supervision, including head movement.

\end{abstract}

\section{Introduction}
\label{sec:introduction}

360$^\circ$ video understanding is critical for providing intelligent systems with omnidirectional perception.
Simultaneous view in every direction helps agents better react in challenging environments for indoor navigation~\cite{anderson2018r2r}, autonomous driving~\cite{caruso2015large,yogamani2019woodscape}, and drone navigation~\cite{caron2018spherical}, to name a few.
Also, virtual reality and 360$^\circ$ action cameras  have pervaded entertainment applications.

Visual saliency prediction is one of the representative benchmarks for 360$^\circ$ video understanding.
It can filter irrelevant or redundant information in panoramic views, and thus promote summarization of 360$^\circ$ videos or dynamic rendering of virtual reality panorama.
Unlike saliency detection in normal field-of-view (NFoV) imagery that often aims to distinguish salient foreground from background, saliency prediction in 360$^\circ$ videos stems from a simple yet nontrivial question: which direction to watch if you were in the scene?
Foreground  objects may not always be of interest, and salient direction is subjective and depends on context.
% For instance, if a person holding an action camera is giving a short tour along the sidewalk, a salient direction should be the one he is pointing at, not the person himself.
Hence, 360$^\circ$ saliency prediction has often been interpreted as automated cinematography~\cite{su2016pano2vid,chou2018self}, highlight detection~\cite{yu2018deep,lee2018memory}, and attention tracking~\cite{hu2017deep360pilot}.

There are a few challenges in predicting visual saliency in 360$^\circ$ videos.
First, in any format of 360$^\circ$ videos (\eg, equirectangular, cubemap~\cite{greene1986environment}), a non-negligible proportion of distortion and discontinuity hinders the accurate processing of omnidirectional view.
Thus, it is nearly impossible to directly leverage models learned from normal videos at no cost.
Some previous works~\cite{lee2018memory,chou2018self,zhu2020auxrn} explicitly project a number of NFoV images from a panorama frame to estimate salient viewpoints.
However, this may not be scalable in terms of both space and time since an order of magnitude larger number of NFoV images (\eg, 81 for \cite{lee2018memory}) should be processed per single panorama frame.
% Although denser projection implies better approximation of saliency map, 
In another line of works, transferrable architectures are proposed to utilize pretrained knowledge from the NFoV domain~\cite{su2017learning,su2019kernel,eder2020tangent}.
They can process 360$^\circ$ input without modification but at the cost of geometric error or additional modules for finetuning.
% Still, \cite{su2019kernel} claimed that layerwise geometry-aware approximation for processing 360$^\circ$ inputs is detrimental to performance for deep convolutional neural networks.

Second, ambiguity is another vital issue for saliency prediction in 360$^\circ$ videos.
% the fundamental question for visual saliency prediction in 360$^\circ$ videos is ``Which direction to watch if you were in the scene?''
While previous works define saliency as intensity and orientation~\cite{itti1998model}, self-information~\cite{bruce2005saliency}, and anomaly~\cite{wang2011image}, there is no definitive answer for which constitutes capture-worthiness or saliency in 360$^\circ$ videos.
A widely accepted tool to interpret saliency is class activation maps (CAM)~\cite{zhou2016learning} in both NFoV domain~\cite{zeng2019multi,zeng2019joint,xie2021online,meng2021foreground} and 360$^\circ$ domain~\cite{cheng2018cube}.
Although CAM can readily capture objects in the scene, it depends on the class labels of the reference dataset and is challenging to integrate with self-supervised pretraining that has no labels.
% Also, as in Fig.~\ref{fig:qualitative},  all foreground objects are not capture-worthy in 360$^\circ$ videos. 
As 360$^\circ$ saliency detection has been usually addressed under minimal or zero supervision, some other works resolve ambiguity by leveraging additional information like the coordinates of target objects~\cite{hu2017deep360pilot} or reference NFoV videos of the same topic~\cite{su2016pano2vid,yu2018deep,lee2018memory}. 

To address these issues, we propose a novel framework for 360$^\circ$ video saliency prediction named \textbf{Pa}noramic \textbf{V}ision Transform\textbf{er} (PAVER). It is equipped with two components: a deformation-aware omnidirectional encoder and a consistency-oriented saliency map decoder.
First, our encoder adopts deformable convolution~\cite{dai2017deformable} to represent a 360$^\circ$ video as a set of small patches with local tangent projection for minimal geometric error. It can replace the NFoV projection that previous works often use with 60$\times$ less geometric error at negligible computation overhead.
Then, we use the Vision Transformer~\cite{dosovitskiy2020image} 
to remove the need for additional finetuning to transfer pretrained weights from the NFoV domain. %with projected patches from deformable convolution as input
As a result, the geometric approximation happens \textit{only once} in our framework, unlike previous deep CNN-based approaches that perform at every layer, relieving the model of layerwise geometric error accumulation~\cite{su2019kernel}.
Our work is the first attempt to exploit the vision transformer to process 360$^\circ$ imagery.

% Our approach does not depend on a specific feature encoder or training process, provided the model is based on the Vision Transformer.
Second, for our decoder to determine capture-worthy context on panoramic videos without supervision or additional information, 
% we interpret saliency in a semantically plausible perspective to tackle ambiguity in 360$^\circ$ video saliency.
we decompose the saliency into three relative relationships of the local patch features from its surrounding contexts.
If the context of a local patch diverges from the overall representation of the video (\textit{local saliency}), the patch can be deemed anomalous and is usually worth noticing.
Moreover, if the spatial and temporal neighbors of a patch are capture-worthy (\textit{spatial \& temporal saliency}), the patch should also be capture-worthy.
% Instead of directly predicting scalar saliency scores, 
By enforcing this simple yet straightforward objective in the feature dimension, we outperform the previous state-of-the-art model by 23\% in the Wild360 dataset~\cite{cheng2018cube}.% without additional labels.
Also, we leverage this saliency prediction for omnidirectional video quality assessment for virtual reality (VR) in VQA-ODV \cite{li2018bridge}, which is crucial for user experience in VR.

In conclusion, we summarize our main contribution as follows.
\begin{enumerate}
  \item Our PAVER  framework is the first attempt to adopt the Vision Transformer~\cite{dosovitskiy2020image} to encode the omnidirectional imagery.
  Along with deformable convolution~\cite{dai2017deformable}, our encoder alleviates geometric projection errors with no additional module and trivially processes panoramic videos in various formats by transferring the weights learned from the normal video datasets.
 \item Thanks to our powerful encoder, we demonstrate that it is sufficient for 360$^\circ$ video saliency prediction to simply learn from relative relations among local patch features,
  outperforming state-of-the-art models for the Wild360 benchmark~\cite{cheng2018cube} by large margins with no additional annotations.
  \item For the applicability of PAVER, we show that PAVER can consistently improve the performance of omnidirectional video quality assessment in the VQA-ODV~\cite{li2018bridge} benchmark with no human supervision like head movement.
    % \item We outperform the state-of-the-art 360$^\circ$ video saliency prediction models for the Wild360 benchmarkt~\cite{cheng2018cube} by a large margin and achieve comparable performance on the omnidirectional video quality assessment benchmark in VQA-ODV \cite{li2018bridge} without human supervision.
\end{enumerate}

\section{Related Work}
\label{sec:related_work}

\textbf{Panoramic Video Processing.}
% There has been a surge of interest in processing data with spherical geometry from molecular structure and climate modeling to omnidirectional cameras.
Efficient and accurate processing of 360$^{\circ}$ images or videos has been studied much. %for various applications, including embodied navigation~\cite{zhu2020auxrn}, video summarization~\cite{lee2018memory}, autonomous driving~\cite{caruso2015large,yogamani2019woodscape}, etc.
% One of the most straightforward methods is to regard panoramic videos as normal field-of-view (NFoV) videos at the cost of accuracy.
One of the most popular approaches is to project panorama into a set of normal field-of-view (NFoV) videos.
Despite its simplicity, it has been effective in various tasks like vision-and-language navigation~\cite{zhu2020auxrn}, language-guided view grounding~\cite{chou2018self}, and 360$^\circ$ video summarization~\cite{lee2018memory}.
However, it requires explicit projection of up to 81 NFoV images per panoramic frame, which is less scalable in both space and time.

For the processing of spherical inputs, some prior works suggest designated architectures with spherical correlation~\cite{cohen2018spherical}, spherical convolution with spectral smoothness~\cite{esteves2018learning}, or operations on unstructured grids~\cite{jiang2018spherical}.
Although they ensure favorable mathematical properties like rotational equivariance, they cannot be transferred from model weights trained with large sets of normal images or videos.
Another compelling direction is to use transferrable architectures combined with geometric adaptation modules with finetuning~\cite{su2017learning,su2019kernel}, polyhedral approximation~\cite{lee2019spherephd}, or both~\cite{zhang2019orientation,eder2020tangent}.
However, additional modules for geometric alignment may be detrimental to latency, either in the training or inference step.

We use a transformer architecture where geometric approximation happens only once at the beginning, unlike previous approaches that perform in every layer.
Pretrained weights from the NFoV domain are transferrable to our approach without pretraining for geometric adaptation.
In addition, our approach is format-independent;
not only the equirectangular format, but our model can also compute the cubemap or other formats without explicit conversion. 

\noindent
\textbf{Visual Saliency Detection.}
Saliency detection has been a longstanding problem in computer vision research.
In order to identify visual saliency, previous approaches utilize intensity and orientation with respect to stimulus~\cite{itti1998model}, self-information maximization~\cite{bruce2005saliency}, intrinsic and extrinsic anomaly~\cite{wang2011image}, and self-resemblance~\cite{seo2009nonparametric}.
More recent work focuses on class activation maps~\cite{zhou2016learning}, where high activation value of a certain class implies saliency.
CAM-based methods are widely accepted for multi-source saliency detection~\cite{zeng2019multi}, weakly supervised semantic segmentation~\cite{zeng2019joint}, and object localization~\cite{xie2021online,meng2021foreground}.
We point our readers to a survey of visual saliency detection \cite{ullah2020brief} for further details.

On the other hand, saliency prediction in 360$^\circ$ videos needs to identify capture-worthy viewpoints within the omnidirectional surroundings.
Given a number of possible candidate viewpoints, it aims at providing plausible viewpoints or a heatmap as if someone is watching the scene.
Thus, the saliency in 360$^\circ$ videos is often ambiguous and depends on subjective context.
To resolve this, some works exploit NFoV videos as exemplars of capture-worthiness~\cite{su2016pano2vid,lee2018memory,yu2018deep}.
Other works leverage human supervision of saliency maps \cite{zhang2018saliency} or object tracking information~\cite{hu2017deep360pilot} for training.
Some recent works also utilize learned class activation maps~\cite{cheng2018cube} or natural language narratives~\cite{chou2018self}.
Our approach is also in line with \cite{cheng2018cube,chou2018self} in that we do not require explicit supervision for training.
One key difference is that we do not rely on additional information like CAM or narratives for training.
Instead, we enforce local saliency and spatiotemporal saliency in the feature map of the local patch context.

\noindent
\textbf{Vision Transformers.}
% Inspired by the success of self-attention architectures~\cite{vaswani2017attention},
Vision Transformers~\cite{dosovitskiy2020image} have been drawing much attention for large-scale visual understanding since they reported impressive performance in image classification~\cite{yuan2021tokens,touvron2021going}, object detection and semantic segmentation ~\cite{liu2021swin,xie2021segformer}.
Recently, vision transformers have been adapted to broader domains, including point cloud~\cite{zhao2021point} and video understanding~\cite{gberta2021timesformer,zhang2021vidtr}.
Another line of works focuses on transferring pretrained knowledge of vision transformers in an unsupervised or semi-supervised manner by leveraging self-distillation~\cite{caron2021emerging}, semantics reallocation~\cite{gao2021tscam}, seed propagation \cite{simeoni2021lost}, and normalized cut \cite{wang2022tokencut}.

Our work is the first to adopt the vision transformer to process omnidirectional inputs.
Closest to our approach is Yun \etal~\cite{yun2021improving}, which utilizes the transformer for indoor semantic segmentation with monocular 360$^\circ$ images.
However, Yun \etal do not take into account 360$^\circ$ format when processing images but instead discard the near-polar region where the spherical distortion is severe.
On the other hand, our approach can process the whole panorama in a format-aware manner without discarding any parts, while being applicable to both 360$^\circ$ images and videos with only trivial overhead.

\section{Approach}
\label{sec:approach}

%------------------------------------------------------------------------------
% Figure 1: Key Idea
\begin{figure}[t]
\centering
\includegraphics[trim=0.0cm 0.0cm 0cm 0.0cm,clip,width=\textwidth]{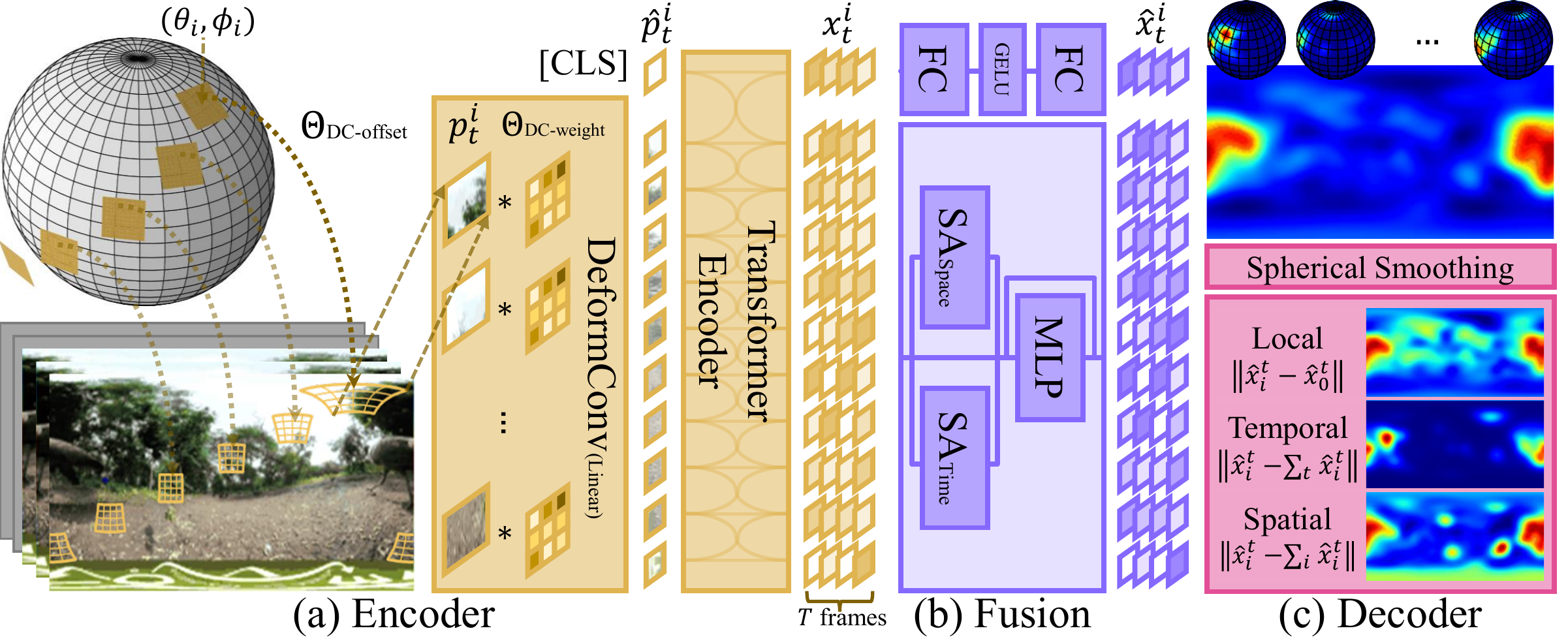}
%\vspace{-5pt}
\caption{Model architecture of \textbf{PAVER} (Panoramic Vision Transformer).}
%\vspace{-10pt}
\label{fig:architecture}
\end{figure}
%------------------------------------------------------------------------------

%\subsection{Preliminary}
% \label{sec:approach_prelim}

We present a model named \textbf{Pa}noramic \textbf{V}ision Transform\textbf{er} (PAVER) for saliency detection in 360$^\circ$ videos, as illustrated in Fig.~\ref{fig:architecture}.
Given a 360$^\circ$ input video in equirectangular format $V=\{v^t\}_{t=1}^T \in \mathbb{R}^{T \times 3 \times H \times W}$ with $T$ frames, our objective is to compute their saliency maps, \ie, $\{\hat{y}^t\}_{t=1}^{T} \in \mathbb{R}^{T \times H \times W}$.
In Sec.~\ref{sec:approach_encoder}, we explain the Transformer encoder for encoding local patches with minimal geometric error (Fig.~\ref{fig:architecture}-(a)).
Then, we deal with the spatiotemporal fusion module for learning consistent local and global features in Sec.~\ref{sec:approach_fusion} (Fig.~\ref{fig:architecture}-(b)).
Finally, we account for the saliency map decoder with spherical smoothing % (Fig.~\ref{fig:architecture}-(c))
and learning objectives in Sec.~\ref{sec:approach_decoder}--\ref{sec:approach_objective}.
Commonly used variables are described as follows:
%\newline
%\vspace{-10pt}

\noindent
\begin{minipage}{\textwidth}
\begin{minipage}[b]{\textwidth}
%\begin{table}[h!]
\centering
%\vspace{-12pt}
\begin{tabular}{c|l}
\hline
$S$    & Length of a tangential patch \\
$W, H$ & Resolution of 360$^\circ$ video input (\eg, $W=448, H=224$) \\
$w, h$ & Number of patches along width and height, \ie, $w=\frac{W}{S}, h=\frac{H}{S}$ \\
$N$    & Number of flattened patches, \ie, $N=w \times h$ \\
$C$    & Number of channels per feature (\eg, 768 for ViT-B/16~\cite{dosovitskiy2020image}) \\ \hline
\end{tabular}
%\end{table}
%\vspace{-25pt}
\end{minipage}
\end{minipage}

\subsection{The Encoder for 360$^\circ$ Videos}
\label{sec:approach_encoder}

\textbf{Local Patch Projection.}
Na{\"i}ve projection of local patches in vision transformers~\cite{dosovitskiy2020image} cannot handle distortion and discontinuity in 360 domains.
Hence, we leverage deformable convolution~\cite{dai2017deformable}, which can process freeform deformation of convolution kernels with marginal computation overhead. %  using learnable offsets.
For each 360$^\circ$ frame $v^t \in \mathbb{R}^{3 \times H \times W}$, we first compute tangential patches with a size of $S \times S$, namely $\{p^t_i\}_{i=1}^{N}$ where $N=wh=\frac{W}{S} \times \frac{H}{S}$.
Then, we linearly project patches to obtain $\{\hat{p}^t_i\}_{i=1}^{N}$ where $\hat{p}^t_i \in \mathbb{R}^{C}$, using deformable convolution with fixed offsets: % to perform both operations at once:
% Instead of explicitly projecting tangential patches, we 
\begin{align}
\label{eq:deformconv}
    \hat{p}^t &= \text{Conv}(p^t) =  \text{DeformConv}(v^t;\Theta_{\text{DC-weight}},\Theta_{\text{DC-offset}}), \\
    \Theta_{\text{DC-offset}}(\theta_i, \phi_i) &= f_{\text{Sph}\rightarrow\text{ER}}(f_{\text{3D}\rightarrow\text{Sph}}(\mathcal{P} \times R(\theta_i, \phi_i) / \lVert \mathcal{P} \times R(\theta_i, \phi_i) \rVert_2)), \\
    R(\theta_i, \phi_i) &=
    \begin{pmatrix}
    \cos\phi_i \cos\theta_i & -\cos\phi_i \sin\theta_i & \sin\theta_i \\
    \sin\theta_i & \cos\theta_i & 0 \\
    \sin\phi_i \cos\theta_i & -\sin\phi_i \sin\theta_i & \cos\phi_i
    \end{pmatrix},
\end{align}
\noindent where $(\theta_i, \phi_i) \in (0,2\pi)\times(-\pi/2,\pi/2)$ is the longitude and latitude of the center of the $i$-th patch, and $\mathcal{P} \in \mathbb{R}^{S\times S\times 3}$ indicates sampled 3D points from an $S \times S$ patch on $z=1$ plane.
$f_{\text{3D}\rightarrow\text{Sph}}$ is the conversion from the 3D cartesian coordinates to the spherical coordinates, while $f_{\text{Sph}\rightarrow\text{ER}}$ is that from the spherical to the 2D equirectangular coordinates.
In a nutshell, we compute the offset for DeformConv by rotating the reference patch $\mathcal{P}$ by $\theta_i$ and $\phi_i$ and projecting it onto an equirectangular plane. 
Please refer to the Appendix for more details.

In Eq.~\ref{eq:deformconv}, the weights for linear projection $\Theta_{\text{DC-weight}}$ can be transferred from pretrained Vision Transformers \cite{dosovitskiy2020image,touvron2021going,caron2021emerging,chen2021empirical} without additional tuning, as $\Theta_{\text{DC-offset}}$ locally projects curved surfaces to flat NFoV images.
Unlike deformable convolutional networks~\cite{dai2017deformable}, the offset $\Theta_{\text{DC-offset}}$ is computed only once and fixed throughout the training.
That is, throughout our architecture, we apply geometric approximation only here for local tangential patches.
Another benefit is that we can adapt any format other than equirectangular (\eg, cubemap) without an explicit conversion process as long as the offset $\Theta_{\text{DC-offset}}$ is computed from the conversion formula between different formats.
Considering that recently more videos are using the equiangular cubemap format for better resolution, our approach can represent the video as-is, regardless of the formats.

\textbf{Transformer Encoder}.
We prefix $N$ patches $\{\hat{p}^t_i\}_{i=1}^{N}$ with a learnable vector $\hat{p}^t_0$ (\ie, \texttt{[CLS]} token) and feed them to the transformer encoder:
\begin{align}
\label{eq:vit}
    \{x^t_i\}_{i=0}^{N} &= \text{Transformer}(\{\hat{p}^t_i\}_{i=0}^{N};\Theta_{\text{ViT}}),
\end{align}
\noindent 
where $\{x^t_i\}_{i=0}^{N}$ is the output features of the patches.
As in the local patch projection of Eq.~\ref{eq:deformconv}, pretrained weights on the NFoV domain (\eg, ViT~\cite{dosovitskiy2020image} trained with ImageNet-21k~\cite{deng2009imagenet}) can be reused for $\Theta_{\text{ViT}}$ without finetuning or even additional modules.
Unless mentioned otherwise, we use vanilla vision transformers with the ViT-B/16 weights~\cite{dosovitskiy2020image}.
Refer to Table~\ref{tab:analysis}-(b) later for the experiments with other transformer variants.
More details are deferred to Appendix.
% This especially stands out for large-scale panoramic visual processing models, since there are no large-scale dataset or common pretraining scheme available.

\subsection{Spatiotemporal Fusion}
\label{sec:approach_fusion}

Since the encoded features $\{x^t_i\}_{i=0}^{N}$ are obtained from each patch separately, we now train a spatiotemporal fusion module so that the local features are smoothly aligned with respect to space and time.
% In order to update global context independently from local context,
We decouple global context $x^t_0$ from the local patch context $\{x^t_i\}_{i=1}^{N}$ and model them separately.
We observe that training both global and local context with identical weight is detrimental to performance, which is further discussed in Sec.~\ref{sec:exp_sal}.

\textbf{Global Context.}
\texttt{[CLS]} tokens in Transformers are generally used as an input to a classification head to predict the output. % for the target task or pretext tasks.
Thus, we regard $x^t_0 \in \mathbb{R}^C$, the output of transformer for the \texttt{[CLS]} input token, as encapsulating the global context of scenery in a sense.
Finally, we encode global context $\hat{x}^t_0 \in \mathbb{R}^C$ via a simple multilayer perceptron (MLP):
\begin{align}
\label{eq:global}
    \hat{x}^t_0 &= \text{MLP}_{\text{G}}(x^t_0; \Theta_{\text{G}}),
\end{align}
\noindent
where MLP refers to two fully-connected layers with GELU activation~\cite{hendrycks2016gelu}.

\textbf{Local Context.}
To encode  temporal information in the local patch context, we use features from all $T$ frames \ie, $X = \{\{x^t_i\}_{i=1}^{N}\}_{t=1}^{T} \in \mathbb{R}^{T \times N \times C}$.
We extend the vanilla transformer encoder with multi-head self-attention~\cite{vaswani2017attention} temporally: %to temporal dimension:
\begin{align}
\label{eq:local_attn}
    \hat{X}  = X' + \text{MLP}_{\text{L}}(X'; \Theta_{\text{L}}), \mbox{ where } X' = X  + \frac{\text{SA}(X; \Theta_{\text{S}}) + \text{SA}(X^T; \Theta_{\text{T}})^T}{2}.
\end{align}
MLP$_{\text{L}}$  indicates an MLP with residual connection, and SA respectively denotes multi-head self-attention along the temporal axis ($T$) and spatial axis ($N$).
$X^T$ is a transpose between the temporal and spatial axis, \ie, $X^T \in \mathbb{R}^{N \times T \times C}$.
By averaging two multi-head self-attentions, local patches have similar feature representation with their spatial and temporal neighbors altogether.
% split spatial attention and temporal attention y

\subsection{The Decoder for Saliency Map}
\label{sec:approach_decoder}

Using both global and local contexts ($\{\hat{x}^t_0\}_{t=1}^{T}$ and $\{\{\hat{x}^t_i\}_{i=1}^{N}\}_{t=1}^{T}$), we first compute the saliency score of each local patch.
We decompose the saliency into three terms: local, temporal, and spatial saliency. 
Instead of directly optimizing saliency scores, we measure relative relations between each local context feature and its neighbors and optimize them on the feature level.

First, the  local saliency measures how much a local patch deviates from the global context.
If the distance between the local patch and global context is large, it can be deemed as an anomalous patch, which may be worth viewing.
Likewise, temporal and spatial saliency reflects how much a local patch deviates from the temporal or spatial mean of its neighbors.
With more distance between, the context of the patch would more differ from those of its neighborhoods. % the local patch feature and spatial \& temporal mean
Finally, our saliency score is computed as follows, where $\alpha=\beta=\gamma=1$ for simplicity:
\begin{align}
\label{eq:score}
    y^t_i &= \alpha \norm{\hat{x}^t_i-\hat{x}^t_0}_2^2 + \beta \norm{\hat{x}^t_i - \frac{1}{T}\sum_{t=1}^{T} \hat{x}^t_i}_2^2 + \gamma \norm{\hat{x}^t_i - \frac{1}{N}\sum_{i=1}^{N} \hat{x}^t_i}_2^2.
\end{align}
\noindent 

\textbf{Spherical Gaussian Smoothing.}
To upsample from $\mathbb{R}^{h \times w}$ to $\mathbb{R}^{H \times W}$ while observing the spherical structure, we apply spherical Gaussian smoothing~\cite{balaji2015understanding}.
The scalar saliency score $\hat{y}^t$ is obtained as
\begin{align}
\label{eq:sphericalsmoothing}
    \hat{y}^t_{ij}(\theta_i, \phi_i, \psi_j) = y^t_i \times \cos\phi_i \times \frac{a}{\sinh\,a} e^{a\cos\psi_j}, \mbox{ where }
    a = \frac{W^2}{4\pi^2\sigma^2}.
\end{align}
\noindent 
$\psi_j$ denotes how much the $j$-th pixel deviates from $(\theta_i, \phi_i)$, and $\sigma$ is the standard deviation of Gaussian smoothing.
Note that spherical Gaussian smoothing is only applied for evaluation purposes.
% For training, we only use the sparse saliency map $Y=\{y_t^i\}_{i=1}^{N} \in \mathbb{R}^{h \times w}$.

\subsection{Learning Objectives}
\label{sec:approach_objective}

Without any ground truth label or additional information, we train the model by ensuring spatiotemporal consistency while maintaining the global context.

\textbf{Temporal Consistency Loss.}
It is natural for two adjacent frames to display similar saliency values and feature distributions.
We take into account two neighboring frames, \ie, $t+1$ and $t-1$ for the $t$-th frame: 
\begin{align}
\label{eq:temploss}
    \mathcal{L}_{T} = \frac{1}{NT} \sum_{t=1}^{T} \sum_{i=1}^{N} \norm{\hat{x}^t_i - \frac{\hat{x}^{t+1}_i+\hat{x}^{t-1}_i}{2}}_2^2.
\end{align}

\textbf{Spatial Consistency Loss.}
Adjacent patches should retain similar saliency scores and feature distributions.
We use the geodesic distance between patches to reflect the spherical structure. 
\begin{align}
\label{eq:spatloss}
    \mathcal{L}_{S} &= \frac{1}{NT} \sum_{t=1}^{T} \sum_{i=1}^{N} \norm{\hat{x}^t_i - \sum_{j=1}^{N} \frac{\gamma_i \delta_{ij} \hat{x}^t_j}{g_{ij}}}_2^2,
\end{align}
\noindent 
where $g_{ij} = ||(\theta_i,\phi_i) - (\theta_j,\phi_j)||_g$ is the geodesic distance between the $i$-th and $j$-th patch, and $\delta_{ij}=1$ when $0 < g_{ij} < \epsilon$ for some $\epsilon$. $\gamma_i$ is a scaling factor such that $\sum_{j=1}^{N}  \frac{\gamma_i \delta_{ij}}{g_{ij}} = 1$. %  (\ie, $S$)
The idea of Eq.~\ref{eq:spatloss} is that the similarity of patches within a certain threshold of $\epsilon$ should be inversely proportional to their geodesic distance.

\textbf{Global Context Loss.}
To train the MLP$_{\text{G}}$ for global context, we encourage all $T$ frames in a video to have a similar global context:
\begin{align}
\label{eq:regloss}
    \mathcal{L}_{G} = \frac{1}{T} \sum_{t=1}^{T} \norm{\hat{x}^t_0 -  \frac{1}{T} \sum_{s=1}^{T}\hat{x}^{s}_0}_2^2.
\end{align}
\noindent 
To sum up, our loss function is defined as follows:
\begin{align}
\label{eq:loss_total}
    \mathcal{L}_{\text{total}} &= \lambda_T \mathcal{L}_T + \lambda_S \mathcal{L}_S + \lambda_G \mathcal{L}_G.
\end{align}

\textbf{Training}.
%\label{sec:approach_training}
We end-to-end train our model with a batch size of 1 and $T=5$ frames per input.
We fix the encoder weight with the pretrained weight of the Vision Transformer (ViT-B/16)~\cite{dosovitskiy2020image}.
We optimize with Adam optimizer~\cite{kingma2014adam} with a learning rate of 2e-7 for 
five epochs.
For hyperparameter, we use $\lambda_T=20, \lambda_S=0.5, \lambda_G=0.1$.
Please refer to the Appendix for more details.

\section{Experiments}
\label{sec:exp}

For evaluation, we perform experiments on two benchmark tasks with Wild360 \cite{cheng2018cube} and VQA-ODV~\cite{li2018bridge} datasets.
First, we evaluate our target task, saliency prediction on 360$^\circ$ videos, on Wild360~\cite{cheng2018cube} as one of the most popular datasets.
Second, we apply our approach to visual quality assessment on 360$^\circ$ videos  on VQA-ODV~\cite{li2018bridge}.
The quality assessment of capture-worthy viewports is important in omnidirectional videos, and the models for this task usually require annotations from headgears or eye-trackers as well as human supervision of subjective assessment.
If we can replace such expensive annotations with visual saliency maps, omnidirectional video quality assessment can become scalable without human supervision.
We thus evaluate how much PAVER can improve the omnidirectional video quality assessment with no such annotations.

\subsection{Experiment Setting}
\label{sec:exp_data}

\textbf{Datasets.} 
Wild360~\cite{cheng2018cube} is composed of 85 video clips about natural scenery, and split into 60 clips for training and 25 for test.
Human annotated ground-truth saliency heatmaps are provided only for test split.
VQA-ODV~\cite{li2018bridge} consists of 540 impaired videos from 60 lossless reference videos.
Nine types of impairment are applied to each reference video with varying degrees of compression levels and projection types (ERP, RCMP, TSP).
The quality of each impaired video is scored by 20 subjects wearing a head-mounted display, and annotated with the head movement (HM) and eye movement (EM) of a subject.
% As the number of viewports an observer can pay attention to is limited, 

\textbf{Evaluation Metrics.}
We use the standard measures of the two benchmarks.
For Wild360, we report three metrics for saliency detection: AUC-Judd~\cite{riche2013saliency}, AUC-Borji~\cite{borji2013analysis}, and Linear Correlation Coefficient (CC).
AUC-Judd computes the true positive and false positive rate of the saliency map.
AUC-Borji randomly samples pixels to calculate false positive rates of these pixels.
CC measures the linear relationship between the proposed saliency map and ground truth.
We regard CC as the most important metric as recommended by \cite{bylinskii2018different}. 
Please refer to \cite{bylinskii2015saliency} for more details.

For VQA-ODV, differential mean opinion score (DMOS) quantifies the quality of omnidirectional videos perceived by the viewers.
Common objective metrics for visual quality assessment like structural similarity (SSIM)~\cite{wang2004image} and peak signal-to-noise ratio (PSNR) are not necessarily proportional to actual human perception. Hence, Pearson correlation coefficient (PCC), Spearman rank correlation coefficient (SRCC), root mean squared error (RMSE) and mean absolute error (MAE) between the DMOS and target metrics are utilized as the quantitative metrics, which we report.

\textbf{Baselines.}
First, we compare our PAVER model against competitive baselines for predicting 360$^\circ$ video saliency based on optical flow~\cite{weinzaepfel2013deepflow}, gradient flow~\cite{wang2015consistent}, generative adversarial networks~\cite{pan2017salgan}, and class activation map with optical flow~\cite{cheng2018cube}.
We also report the performance of unsupervised saliency detection and unsupervised object discovery models based on vision transformers, including TS-CAM~\cite{gao2021tscam}, DINO~\cite{caron2021emerging}, LOST~\cite{simeoni2021lost} and TokenCut~\cite{wang2022tokencut}.
%(i) TS-CAM~\cite{gao2021tscam} that performs semantics reallocation of class activation maps, (ii) DINO~\cite{caron2021emerging} that uses the attention maps of self-distilled models,
%(iii) LOST~\cite{simeoni2021lost} that relies on the seed propagation of inverse degree attentions, and TokenCut~\cite{wang2022tokencut} that exploits the normalized cut of feature eigenvectors.
For a fair comparison with our approach, we use the identical local patch projection module in Sec.~\ref{sec:approach_encoder} and the spherical Gaussian smoothing module in Sec.~\ref{sec:approach_decoder}. % to process 360$^\circ$ format.

For the ablation study, we also report five variants of our approach.
PAVER (Cartesian) replaces all 360$^\circ$ aware components in our model with normal field-of-view (NFoV) equivalents.
PAVER (NoGlobal) is the model without the MLP$_{\text{G}}$ projection of global context.
PAVER (NoLocal) replaces the local context spatiotemporal fusion module with a simple MLP.
PAVER (NoDecoupled) encodes both global and local contexts together in a single transformer encoder.
PAVER (ScoreLoss) is trained with the spatiotemporal score consistency (\ie, the sparse saliency map $Y$) instead of the feature map consistency.

For VQA-ODV, we report the performance of PSNR, WS-PSNR~\cite{sun2017weighted}, and S-PSNR~\cite{yu2015framework} with different weight conditions: uniform, random, our saliency map, and reversed saliency map.
Here the primary comparison is PSNR variants between our saliency map and the ones with human head movement supervision 
since PSNR metrics weighted with human head and eye movement supervision are considerably better in quality assessment~\cite{li2018bridge}.
More details of baseline models' configuration can be found in Appendix.

\noindent
\begin{minipage}{\textwidth}
\begin{minipage}[b]{0.56\textwidth}
% \begin{table}[!h]

%%%% Wild360 state-of-the-art & ablation
\centering
\captionof{table}{Comparison of saliency prediction accuracy on the Wild360 dataset~\cite{cheng2018cube}.}   
\begin{tabular}{l|ccc}
\hline
 & CC & AUC-J & AUC-B \\ \hline
Motion Magnitude~\cite{weinzaepfel2013deepflow} & 0.288 & 0.687 & 0.642 \\
ConsistentVideoSal~\cite{wang2015consistent} & 0.085 & 0.547 & 0.532 \\
SalGAN~\cite{pan2017salgan} & 0.312 & 0.717 & 0.692 \\
Equirectangular~\cite{cheng2018cube} & 0.337 & 0.839 & 0.783 \\
CubePad (Static)~\cite{cheng2018cube} & 0.448 & 0.881 & 0.852 \\
CubePad (CLSTM)~\cite{cheng2018cube} & 0.420 & \underline{0.898} & \underline{0.859} \\ \hline
TS-CAM~\cite{gao2021tscam}   & 0.414 & 0.831 & 0.802 \\
DINO~\cite{caron2021emerging} & 0.406 & 0.850 & 0.831 \\
LOST~\cite{simeoni2021lost}   & 0.444 & 0.809 & 0.786 \\
TokenCut~\cite{wang2022tokencut} & \underline{0.500} & 0.841 & 0.815 \\\hline
PAVER (NoGlobal)       & 0.376 & 0.814 & 0.797 \\ % 0.7968 0.8138 0.3760 0.3566
PAVER (NoDecoupled)    & 0.492 & 0.881 & 0.860 \\
PAVER (Cartesian)      & 0.549 & 0.898 & 0.875 \\
PAVER (NoLocal)        & 0.561 & 0.895 & 0.873 \\ % 0.8729 0.8955 0.5610 0.4165
PAVER (ScoreLoss)      & 0.575 & 0.906 & 0.883 \\ \hline % 0.8830 0.9060 0.5754 0.4121
\textbf{PAVER}         & \textbf{0.616} & \textbf{0.923} & \textbf{0.899} \\ \hline % 0.8889 0.9118 0.5922 0.4193
\end{tabular}
% \vspace{3pt}
\label{tab:wild360}
% \end{table}
\end{minipage}
\hfill
\begin{minipage}[b]{0.38\textwidth}
    \centering
    \includegraphics[trim=0.0cm 0.0cm 0cm 0.0cm,clip,width=\textwidth]{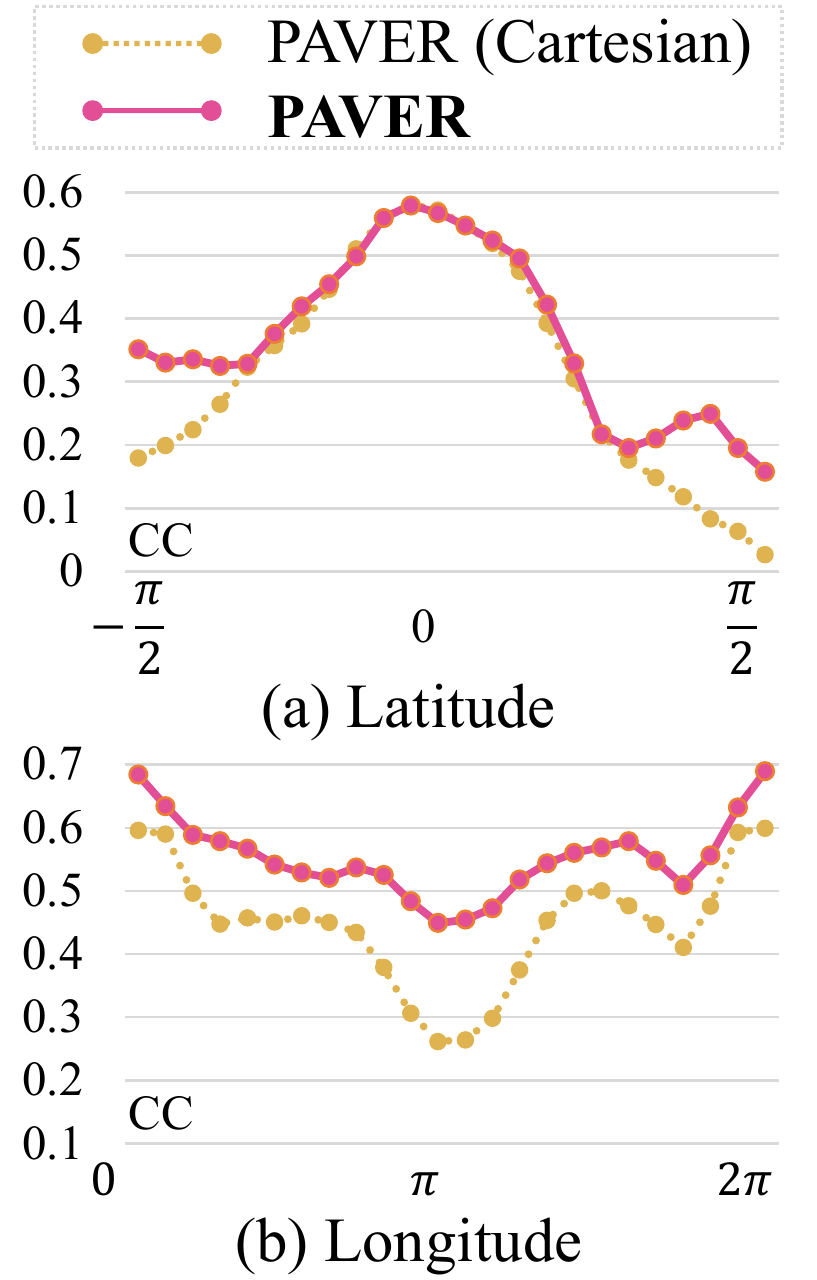}
    %\vspace{-20pt}
    \captionof{figure}{Influence of distortion on Wild360 test split.}
    \label{fig:distortion_exp}
\end{minipage}
\end{minipage}
%\vspace{1em}

\subsection{Results on Saliency Detection}
\label{sec:exp_sal}

\textbf{Comparison with the Prior Arts.}
Table~\ref{tab:wild360} compares the performance with prior arts on the Wild360 dataset.
PAVER outperforms previous state-of-the-art models by large margins: +11.6\%p (\ie, 23\%) in CC, +2.5\%p in AUC-J, and +4.0\%p in AUC-B.
We achieve notable improvement especially in correlation coefficient (CC). %, without using auxiliary information like human viewpoint supervision or class activation maps.
That is, evaluating the relative relations among local patches is sufficient to achieve competent performance in the Wild360 dataset.
Our approach is also time-efficient in that we do not require time-consuming computation for additional information like optical flow; for example, CubePad~\cite{cheng2018cube} requires 19 seconds per frame pair for flow computation~\cite{weinzaepfel2013deepflow}.

Moreover, PAVER consistently achieves better performance compared to the baselines that use the vision transformer as the encoder backbone (\ie, TS-CAM,  DINO, LOST, and TokenCut).
It is worth noticing that merely applying pretrained weights of the vision transformers does not guarantee superior performance.
For instance, TS-CAM or DINO show slightly worse performance than the best performing non-ViT model.
Although they are competent in detecting larger foreground objects, 360$^\circ$ videos usually contain multiple small objects of potential interest.
As in Fig.~\ref{fig:qual_wild360}-(a), the baselines with transformer encoders relatively fall short when multiple salient objects are worth viewing in the scene.

\textbf{Ablation Studies.}
The last six rows in Table~\ref{tab:wild360} compare the PAVER variants.
First of all, performance drops significantly when the global context is poor, as PAVER without the global context encoder (\ie, MLP$_\text{G}$) plummets by 24\%p in CC.
PAVER without the decoupled global-local context encoder decreases CC by 12.4\%p, which suggests that independent parameter update for the global context encoder is essential for better performance.
The performance of PAVER without 360$^\circ$ geometry-aware modules drops by 6.7\%p, implying the importance of the 360$^\circ$ format awareness.
Also, as in Fig.~\ref{fig:distortion_exp}-(a), the gap between PAVER (Cartesian) and PAVER particularly widens in the near-polar region (\eg, $\times 4$ for $\theta=\frac{\pi}{2}$), where most of the distortion in 360$^\circ$ videos takes part.
Replacing the spatiotemporal local context encoder with a simple MLP decreases CC by 5.5\%p.
The residual connections of both spatial and temporal self-attention are beneficial for generating saliency maps with high fidelity.
If we enforce score-level consistency instead of feature-level consistency during training, the performance drops by 4.1\%p in CC.
In summary, all components in PAVER contribute to the full model's performance in their own ways.

\begin{table}[t]
\centering
\caption{(a) Performance comparison with different saliency score compositions, where L, T, and S denotes Local, Temporal, and Spatial saliency score, respectively. (b) Influence of different backbone architectures, pretrained weights, and resolution in our encoder on the Wild360 Dataset~\cite{cheng2018cube}.}
\begin{tabular}{c@{\hskip 10pt}c}
\centering
%%%% Score decomposition
\begin{tabular}{ccc|ccc}
\multicolumn{6}{c}{(a)} \\
\hline
L & T & S         & CC    & AUC-J & AUC-B \\ \hline
 & \checkmark &   & 0.276 & 0.760 & 0.731 \\
\checkmark &  &   & 0.548 & 0.897 & 0.875 \\
 &  & \checkmark  & 0.552 & 0.894 & 0.872 \\
 & \checkmark & \checkmark & 0.557 & 0.899 & 0.876 \\
\checkmark &  & \checkmark & 0.585 & 0.907 & 0.884 \\ \hline
\checkmark & \checkmark & \checkmark  & \textbf{0.616} & \textbf{0.923} & \textbf{0.899} \\ \hline
% \multicolumn{6}{c}{(a)}
\end{tabular}
&
%%%% Different backbones
\begin{tabular}{l|c|ccc}
\multicolumn{5}{c}{(b)} \\
\hline
Backbone   & Res. & CC & AUC-J & AUC-B \\ \hline
TimeSformer (T=16)~\cite{gberta2021timesformer} & 224 & 0.465 & 0.863 & 0.843 \\ % (T=16)
DINO (B/16)~\cite{caron2021emerging} & 224 & 0.524 & 0.893 & 0.871 \\ 
DINO (B/8)~\cite{caron2021emerging}  & 224 & 0.563 & 0.905 & 0.881 \\
ViT-Ti/16~\cite{touvron2021training} & 224 & 0.540 & 0.904 & 0.881 \\ 
ViT-B/16~\cite{dosovitskiy2020image} & 384 & 0.567 & 0.893 & 0.871 \\ 
% ViT-B/16~\cite{dosovitskiy2020image} & 448 & & & \\
\hline
ViT-B/16~\cite{dosovitskiy2020image} & 224 & \textbf{0.616} & \textbf{0.923} & \textbf{0.899} \\ \hline
% \multicolumn{5}{c}{(b)}
\end{tabular}

\end{tabular}
\label{tab:analysis}
%\vspace{-12pt}
\end{table}

\textbf{Analysis on Saliency Score Composition.}
Table~\ref{tab:analysis}-(a) summarizes the influence of different saliency score components in Eq.~\ref{eq:score}.
Computing saliency maps only with temporal saliency does not show acceptable performance.
On the other hand, when the temporal saliency score is added to other sets of scores (\ie, S$\rightarrow$T$+$S, L$+$S$\rightarrow$L$+$T$+$S), the performance consistently improves in all three metrics.
Since the deviation of a local patch along the timeline is relatively small in magnitude, temporal saliency helps our saliency prediction update smoothly in time.
Using only local or spatial saliency displays similar performance, but both of them fall short by 6--7\%p when compared to the full model.
These two saliency scores are complementary in that combining local and spatial saliency boosts performance by 3--4\%p in CC.
% Qualitative aspects of different saliency scores can be found in Fig.~\ref{fig:qual_wild360}-(b-c).

\textbf{Influence of Encoder Backbones.}
Table~\ref{tab:analysis}-(b) compares the performance of our model with different architectures, pretrained weights, and resolutions.
First, we replace the video transformer in PAVER with TimeSformer~\cite{gberta2021timesformer}. It shows 
inferior performance mainly because the model is trained with a larger temporal hop size, as capturing subtle movement in scenery is essential for better saliency maps in the Wild360 dataset.
When we use DINO~\cite{caron2021emerging}, which requires even no labels as the backbone of our model,
it shows better performance compared to the SOTA models.
The model with a smaller patch size reports better CC by 4\%p, which is presumably due to higher fidelity of the saliency map.
Using ViT-Ti/16~\cite{touvron2021training}, the performance drops by 6.6\%p in exchange for 15$\times$ fewer model parameters.
This is still better than existing SOTA models for all three metrics.

\subsection{Omnidirectional Video Quality Assessment}

\begin{table}[t]
\centering
\caption{Results of omnidirectional video quality assessment on VQA-ODV~\cite{li2018bridge}. The better is higher PCC and SRCC or lower RMSE and MAE.}
\resizebox{\columnwidth}{!}{
\begin{tabular}{l|cccc|cccc|cccc}
\hline
       & \multicolumn{4}{c}{PSNR} & \multicolumn{4}{|c}{WS-PSNR~\cite{sun2017weighted}} & \multicolumn{4}{|c}{S-PSNR~\cite{yu2015framework}}\\ 
Weight    & PCC   & SRCC  & RMSE  & MAE   & PCC   & SRCC  & RMSE  & MAE   & PCC   & SRCC  & RMSE  & MAE \\ \hline
None      & 0.650 & 0.664 & 9.004 & 7.027 & 0.672 & 0.684 & 8.771 & 6.909 & 0.693 & 0.698 & 8.541 & 6.681 \\
Random    & 0.650 & 0.663 & 9.004 & 7.027 & 0.672 & 0.684 & 8.771 & 6.909 & 0.693 & 0.698 & 8.540 & 6.680  \\
Reverse   & 0.646 & 0.654 & 9.041 & 6.883 & 0.646 & 0.660 & 9.044 & 7.033 & 0.683 & 0.693 & 8.652 & 6.711 \\
DINO~\cite{caron2021emerging} & 0.657 & 0.677 & 8.934 & 7.105 & 0.674 & 0.690 & 8.747 & 7.033 & 0.699 & 0.710 & 8.468 & 6.665 \\
PAVER     & 0.657 & 0.664 & 8.931 & 7.133 & 0.692 & 0.704 & 8.551 & 6.794 & 0.702 & 0.707 & 8.438 & 6.659 \\ \hline
HM(Supervised)    & 0.733 & 0.726 & 8.054 & 6.479 & 0.731 & 0.722 & 8.086 & 6.565 & 0.736 & 0.741 & 8.022 & 6.305 \\ \hline
\end{tabular}
}
\label{tab:vqaodv}
%\vspace{-5pt}
\end{table}

Table~\ref{tab:vqaodv} reports performance metrics between PSNR variants and DMOS.
When using random weights for PSNR computation, the results are nearly identical to the ones with no weight assignment.
That is, providing weights that are irrelevant to saliency does not improve the performance.
If we use saliency maps from the PAVER as PSNR weights, the performance consistently improves, \eg, 0.7\%p in PSNR, 2.0\%p in WS-PSNR, and 0.9\%p in S-PSNR, respectively for the PCC metric.
On the other hand, if we use reverse saliency maps as weights, the performance worsens compared to the PSNR variants without weight assignment.
For instance, using correct saliency maps and reversed saliency maps displays a 4.6\%p gap for the PCC metric in WS-PSNR.
This implies that proper assignment of saliency maps helps solve the omnidirectional quality assessment.

\textbf{Comparison with Head Movement Supervision.}
The last row of Table~\ref{tab:vqaodv} reports the performance of PSNR variants with actual human head movement as weights.
Compared to S-PSNR with head movement supervision, S-PSNR with our saliency maps shows 3.4\%p lower performance in PCC.
Unlike ground truth labels that require trackable headgears for annotation, our saliency map can be automatically computed using a couple of videos.
%Still, S-PSNR weighted with our saliency map shows 5.2\%p better performance than vanilla PSNR, which implies the feasibility of omnidirectional video quality assessment without human supervision.

\subsection{Qualitative Results.}
\label{sec:exp_quant}

\begin{figure}[t]
\centering
\includegraphics[trim=0.0cm 0.0cm 0cm 0.0cm,clip,width=\textwidth]{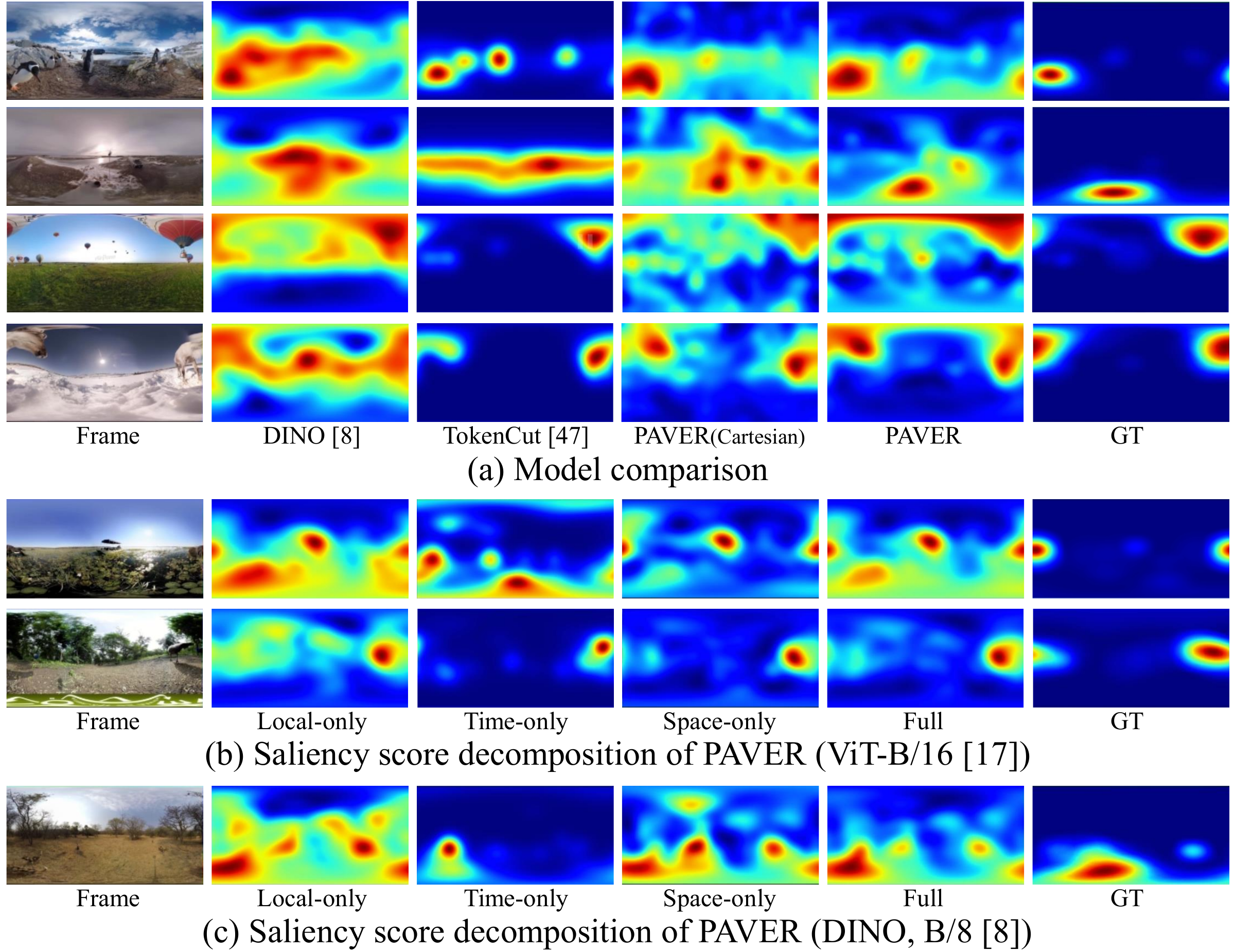}
%\vspace{-12pt}
\caption{Qualitative comparison of saliency prediction on Wild360~\cite{cheng2018cube}.}
\label{fig:qual_wild360}
\end{figure}

\textbf{Saliency Detection.}
Fig.~\ref{fig:qual_wild360}-(a) compares some examples of saliency prediction by different methods.
In general, our PAVER better captures contexts that are worth viewing.
For example, our model places the highest score on the penguin moving towards the camera (a-1) or the polar bear instead of a jeep or a flag nearby (a-2). 
Also, compared to PAVER(Cartesian), our model can propose more sphere-aware saliency maps as in the second and third rows of  Fig.~\ref{fig:qual_wild360}-(a).

Fig~\ref{fig:qual_wild360}-(b-c) compares the PAVER results according to the score decompositions.
Local-only and space-only saliency maps look alike in that they both assign higher scores on anomalous patches.
However, unlike space-only saliency maps, local-only saliency maps tend to favor object-like patches for both supervised~\cite{dosovitskiy2020image} and self-supervised pretraining~\cite{caron2021emerging}. 
Time-only saliency focuses on subtle movement in the scene, which helps generate smooth transitions of saliency maps.
We present more qualitative examples in Appendix.

%------------------------------------------------------------------------------
% Figure 3: Qualitative - Wild360
%------------------------------------------------------------------------------

\begin{figure}[t]
\centering
\includegraphics[trim=0.0cm 0.0cm 0cm 0.0cm,clip,width=\textwidth]{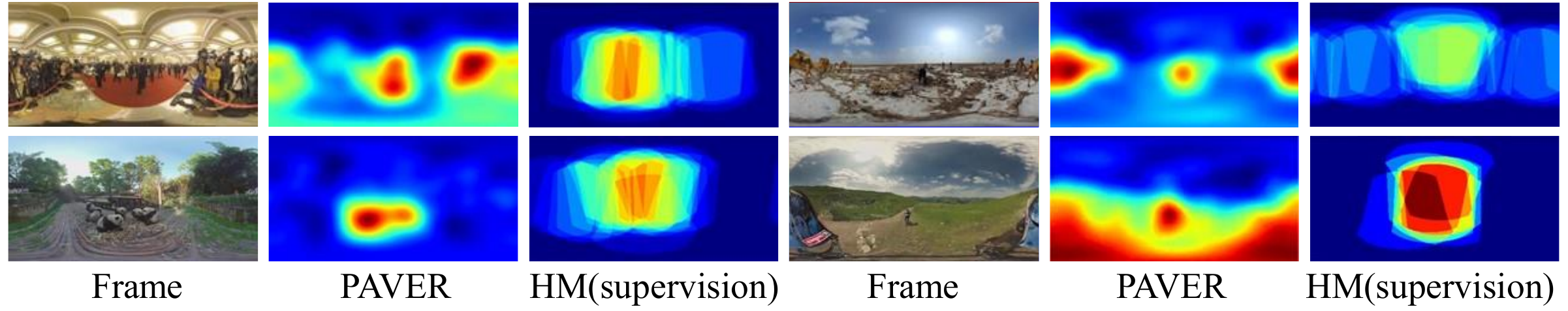}
%\vspace{-12pt}
\caption{Comparison of saliency and video quality prediction on VQA-ODV~\cite{li2018bridge}.}
\label{fig:qual_vqaodv}
%\vspace{5pt}
\end{figure}

\textbf{Quality Assessment.}
Fig.~\ref{fig:qual_vqaodv} illustrates our saliency prediction on the VQA-ODV dataset.
Compared to head movement, our saliency maps are in line with what people think is worth viewing.
Still, as in Fig.~\ref{fig:qual_vqaodv}-(4), our model struggles in cases where the camera drastically moves.

%------------------------------------------------------------------------------
% Figure 4: Qualitative - VQAODV
%------------------------------------------------------------------------------

\begin{figure}[t]
\centering
\includegraphics[trim=0.0cm 0.0cm 0cm 0.0cm,clip,width=\textwidth]{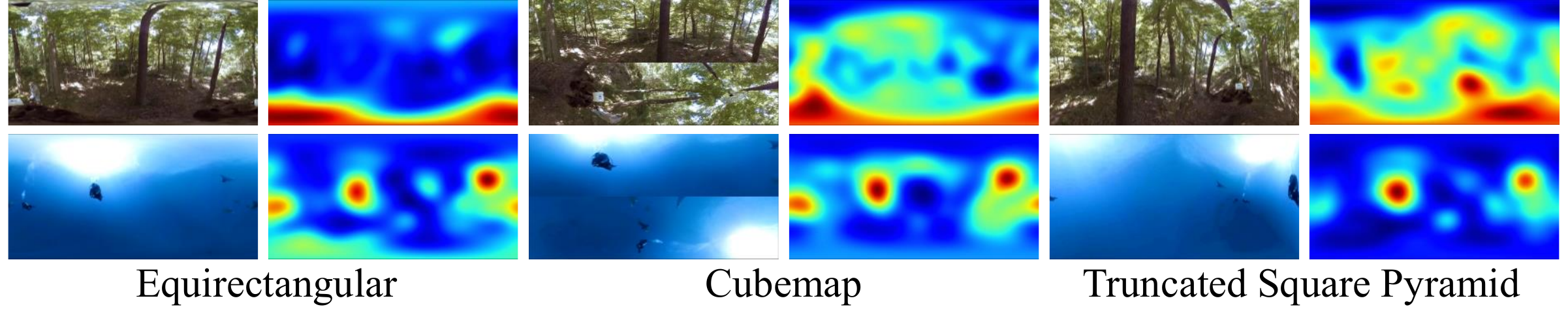}
%\vspace{-12pt}
\caption{Prediction comparison of different projection formats for 360$^\circ$ videos.}
\label{fig:qual_format}
\end{figure}

\textbf{360$^\circ$ Videos with Different Formats.}
Fig.~\ref{fig:qual_format} displays our saliency prediction on varying 360$^\circ$ video projection formats, including equirectangular (ERP), cubemap (RCMP), and truncated square pyramid projection (TSP).
Using randomly sampled videos from the test split of Wild360, we convert them from ERP to RCMP or TSP using 360tools\footnote{https://github.com/Samsung/360tools}.
We replace $\Theta_{\text{DC-offset}}$ for ER with the offsets for each video format.
Computation of $\Theta_{\text{DC-offset}}$ for different video formats can be found in Appendix.
Our model can process different 360$^\circ$ formats without explicitly converting from one to another while returning highly consistent saliency maps regardless of the formats, especially for the ones with severe regional sampling discrepancy like TSP.

%------------------------------------------------------------------------------
% Figure 4: Qualitative - VQAODV
%------------------------------------------------------------------------------

\section{Conclusion}
\label{sec:conclusion}

We proposed a new model for 360$^\circ$ video saliency prediction named PAVER.
We regarded 360$^\circ$ videos as a set of patches using deformable convolution, which alleviates the need for layerwise geometric approximation unlike other CNN-based approaches.
We adopted the vision transformer to encode omnidirectional imagery by reusing pretrained knowledge from normal videos with no need for complex adaptation.
Even with three simple feature-wise consistency objectives, PAVER outperformed prior arts that use additional annotations or vision transformers in Wild360.
We also achieved consistent improvement in omnidirectional video quality assessment in VQA-ODV. % to demonstrate the applicability of our proposed approach.
% We have shown that mere adaptation of ViT does not guarantee performance in Wild360 benchmark, while our saliency score decomposition and geometry-awareness independently contribute to the full model's performance.
%We have demonstrated that decoupled global and local fusion module as well as geometric awareness independently contributes to our performance.
Last but not least, all computations in PAVER can be adapted for various 360$^\circ$ formats without explicit conversion.

There are multiple interesting future directions beyond this work.
First, we can extend PAVER to be adaptable with multi-scale vision transformers such as % Our approach can adapt most of the vision transformers with single scale, but it is not trivial to extend to
Swin transformer~\cite{liu2021swin}, MViT~\cite{fan2021multiscale}, SegFormer~\cite{xie2021segformer}, and SeTR~\cite{SETR}.
Since they can be effective for fine-grained prediction, geometry-aware multi-scale encoding can be beneficial for a better understanding of omnidirectional imagery.
Second, our work could be used as a generic omnidirectional encoder for various tasks like language-guided view grounding~\cite{chou2018self}, embodied navigation~\cite{zhu2020auxrn}, and 360$^\circ$ video question answering~\cite{yun2021panoavqa}.
Another direction is to combine both 360$^\circ$ and NFoV inputs to train a unified architecture to leverage the complementary nature of the two formats.

\textbf{Acknowledgement}. 
We thank Youngjae Yu, Sangho Lee, and Joonil Na for their constructive comments.
This work was supported by AIRS Company in Hyundai Motor Company \& Kia Corporation through HKMC-SNU AI Consortium Fund and Institute of Information \& communications Technology Planning \& Evaluation (IITP) grant funded by the Korea government (MSIT) (No.2019-0-01309, No.2019-0-01082). % 01309 (ETRI) and 01082 (StarLab)
Gunhee Kim is the corresponding author.

\clearpage
\bibliographystyle{splncs04}
\bibliography{eccv22-paver}
\end{document}